# FastContext: an efficient and scalable implementation of the ConText algorithm

Jianlin Shi[1], John F. Hurdle[1]

[1]Department of in Biomedical Informatics, University of Utah, Salt Lake City, UT USA

**ABSTRACT**

**Objective**: To develop and evaluate FastContext, an efficient, scalable implementation of the ConText algorithm suitable for very large-scale clinical natural language processing.

**Background**: The ConText algorithm performs with state-of-art accuracy in detecting the experiencer, negation status, and temporality of concept mentions in clinical narratives. However, the speed limitation of its current implementations hinders its use in big data processing.

**Methods**: We developed FastContext through hashing the ConText's rules, then compared its speed and accuracy with JavaConText and GeneralConText, two widely used Java implementations.

**Results**: FastContext ran two orders of magnitude faster and was less decelerated by rule increase than the other two implementations used in this study for comparison. Additionally, FastContext consistently gained accuracy improvement as the rules increased (the desired outcome of adding new rules), while the other two implementations did not.

**Conclusions**: FastContext is an efficient, scalable implementation of the popular ConText algorithm, suitable for natural language applications on very large clinical corpora.

# INTRODUCTION

Algorithmic processing efficiency grows increasingly important as the size of clinical datasets grow, especially in the era of "Big Data"[1] In the realm of clinical natural language processing (NLP), even small increases in processing throughput are important when handling very large note corpora. An extreme example is the U.S. Veterans Administration national data warehouse, which contains nearly three billion notes; Divita et al. concluded that building concept indexes of the entire note set would take hundreds of years without using multiple servers. On the other hand, shaving off even ten milliseconds per note would save over a hundred days per billion notes.

The utilization of information extraction and retrieval is expected to become one of the future trends of clinical research and practice[2]. With the fast growth of clinical data, the need for faster and more accurate information processing has escalated. To overcome the speed issue, we investigated the details of the information extracting process to identify the bottleneck of the processing pipeline: the context detection, which consumes more than 70% processing time in Divita's report. In our preliminary experiments, the percentage was even higher.

In this paper, we present FastContext, an implementation of the state of the art ConText algorithm. FastContext is designed to accelerate the processing speed of ConText without jeopardizing accuracy.

## BACKGROUND

**Context information**

The context information is a set of modifiers that is associated with a concept. In the clinical domain, the context information typically includes three types of modifiers [3]: negation (whether the target concept exits, not exists, or speculated/uncertain), experiencer (whether the target concept refers to patients or not), and temporality (whether the target concept is currently true, historically true, or hypothetical). In some studies, the speculation/certainty is separated as an independent modifier[4, 5]. These modifiers help shape the subtlety of clinical concepts, which is critical to narrate clinical information.

Negation is the most studied modifier, which refers to the statement about what a patient does not have, such as medical conditions. Previous studies discovered that clinical notes contain a significant number of negative statements[6]. In the corpus [7] used in this study, more than one fifth of the identified disorder concepts were negated. These negations are often clinically meaningful, and can support differential diagnoses and treatment planning. For instance, in an ultrasound report, the note "No atrial septal defect is found" rather than the note "Atrial septal defect is found" would lead to completely different diagnoses and treatment directions.

**Context detectors**

Several rule-based context detectors have been developed and evaluated, such as NegExpander[8], NegFinder[6], NegEx[9] and its descendent ConText[3, 10]. NegExpander identifies the negation words and conjunctions, and asserts the conjunctive noun phrases as negated. However, it cannot adjust the negation scope on relevant semantic clues, such as "although". NegFinder introduces "negation terminators" to overcome this limitation. Nevertheless, it cannot handle pseudo-negation words, such as double negations. NegEx and ConText use "pseudo" triggers to deal with these situations. Goryachev et al.[11] compared four different negation detection methods, including NegEx, NegExpander and other two simple machine learning approaches. The rule based NegEx (F measures: 0.89) and NegExpander (F measures: 0.91) outperformed the other two machine learning approaches (F measures: 0.78, 0.86). Several machine learning based or hybrid systems were also explored[4, 12-16]. Only NegClue[15] (a negation detector) and Cogley, et al.'s system[16] (a temporality and experiencer detector) outperformed the rule-based systems.

Although these reported results suggest that the context detection task has been "solved," their performances drop significantly on different corpora outside the corpus that they were developed[17]. That means applying these solutions in real practice often requires additional tuning. Machine learning based approaches need new annotations to be retrained, and rule-based systems need rule modifications. In general, the latter is less labor intensive but computationally expensive in running time. FastContext is a solution that can take the advantage of rule-based system while remaining computationally efficient and scalable.

## METHOD

### FastContext implementation

Observing that the ConText rules search mostly for word by word matches with a few wildcards, FastContext uses hashing to process rules simultaneously. First, it constructs a nested-map structure from the rules. The top level is a single map, where the keys are the first word in each ConText rule. The value associated with a top-level key is a child map which keys consist of the second words of all the rules starting with that first word. Similarly, the rest of the words in the rules will be included in the subsequent descendent maps. With this data structure, FastContext can process all the rules without looping through them, leading to faster processing speed.

FastContext supports all the three types of context lexical cues (trigger term, pseudo-trigger term and termination term) and the three directional properties (forward, backward and bidirectional) for each modifier. Also, it supports the windows size configuration: the maximum scope of a context cue can take effect. This configuration is helpful to restrict the context scope when sentences are difficult to be segmented, which is common in clinical documents and sentence detectors often yield paragraph-length "sentences." A few examples below illustrate how the ConText rule can be configured.

| can rule out | forward | trigger | negated | 10 |
|---|---|---|---|---|

The rule above means if the phrase "can rule out" is found, this context cue will affect as a negated trigger in forward direction: the words come after it (within 10 token windows size) will be negated.

| although | forward | termination | negated | 30 |
|---|---|---|---|---|

This is a termination rule, it will stop the negation cues on its left side to affect forward --- towards this cue's right side.

| false negative | both | pseudo | negated | 30 |
|---|---|---|---|---|

This is a pseudo-trigger rule. "false negative" is a double negation, which has negation cues but not means negated. So whenever this rule is matched, it will ignore the negation trigger cues "false" or "negative." Since its effect is to overwrite other rules, its own direction and window size do not take any effect, but just to keep the rule format consistent.

### Evaluation

Against the two other popular ConText implementations: JavaConText [7] and GeneralConText [18], we evaluated FastContext, measuring both speed and accuracy. Each of the three implementations was used to process the same dataset with the same hardware setting: dedicated nodes with eight processors of Intel Xeon E5530 2.40G and 2.4G memory on each node. Using the nodes to parallel the process enabled each implementation to finish the evaluation within acceptable time.

*Dataset*

For this evaluation, SemEval 2015 "Analysis of Clinical Text" task dataset [19] was used. This dataset has 431 narrative clinical notes with 19512 manually annotated disorder concepts and corresponding modifiers. SemEval annotations differ slightly from ConText annotations. To minimize the modification of ConText implementations, we treated a SemEval annotation that has "Negation: yes" and "Uncertainty: true" as equivalent to ConText's "Negation: possible." Table 1 shows the distribution of different context modifiers.

Table 1 Data composition of test dataset

|  | Modifiers | Number of test cases | Percent |
|---|---|---|---|
| Negation | Negated | 3621 | 22% |
|  | Affirmed | 11796 | 70% |
|  | Possible | 1333 | 8% |
| Experiencer | Patient | 16536 | 99% |
|  | Other | 214 | 1% |

To prepare the evaluation dataset, the notes were pre-processed by a home-grown tokenizer and sentence segmenter (Jianlin Shi et al. 2016). Tokenized words were rejoined using whitespaces to provide fair input for three implementations. 1743 concepts comprised of disjointed words and 1019 concepts segmented into sentences without context information were excluded, leaving 16750 concepts.

*Speed evaluation*

To assess the baseline, the original rules of JavaConText and GeneralConText were used. To evaluate the speed performance with respect to different rule amount, starting from the initial 409 ConText rules directly derived from JavaConText and kept in the same order, additional 440 home-grown rules were incrementally added to the three programs, at a pace of 50 rules per step (40 rules in the last step and 849 final rules in total). At each step, the added rule was randomly selected from the 440 rules. 200 runs were repeated at every step. The average processing time across different implementations was compared at each step.

*Accuracy evaluation*

To determine accuracy with regard to rule amount, the same rule adding steps as used above were followed. The accuracy was measured through F scores using the following formula.

$$F = \frac{2 \times Precision \times Recall}{Precision + Recall}$$

Because 200 runs at each step were repeated, the average F score of each step was calculated by taking the mean of 200 scores.

Because the ConText algorithm is designed to detect the not "Affirmed", we only calculated the F scores for detecting the other two values "Negated" and "Possible." GeneralConText was not evaluated in "Possible" detection, due to its non-support of "Possible" detection. Similarly, because the ConText algorithm is to find the concepts are not related to "Patient", only the F score for "Other" detection was calculated.

# RESULT

## Speed comparison of three implementations

Table 2 shows the average processing time of three implementations using different rule amount. FastContext shows compelling speed improvement over JavaConText and GeneralConText. Additionally, the processing time of both GeneralContext and JavaConText obviously increases with the rule amount, while FastContext's time slightly increases.

Table 2 Average processing time (in milliseconds) of three implementations in 200 runs

| Number of Rules | GeneralConText | JavaConText | FastContext | Times of speed over (rounded): | |
| --- | --- | --- | --- | --- | --- |
| | | | | GeneralConText | JavaConText |
| original rules | 62615.9 | 20204.4 | | | |
| 409 | 64440.3 | 21091.6 | 110.2 | 580 | 190 |
| 459 | 72559.3 | 24142.3 | 123.3 | 590 | 200 |
| 509 | 78740.0 | 26243.7 | 138.5 | 570 | 190 |
| 559 | 85838.1 | 28606.1 | 130.0 | 660 | 220 |
| 609 | 92847.9 | 30443.2 | 141.1 | 660 | 220 |
| 659 | 98195.4 | 33782.2 | 179.6 | 550 | 190 |
| 709 | 105200.4 | 35795.3 | 134.0 | 790 | 270 |
| 759 | 110018.7 | 38430.1 | 152.8 | 720 | 250 |
| 809 | 117706.6 | 42229.2 | 135.6 | 870 | 310 |
| 849 | 122761.2 | 43490.8 | 158.8 | 770 | 270 |

## Accuracy comparison of three implementations

*Comparing the accuracy of negation detection*

The accuracy of negation detection is reflected by the average F scores of "Negated" detection and "Possible" detection. Figure 2 compares the average F scores of "Negated" detection among the three implementations. The average F score of JavaConText using original rules (0.616) is not plotted to zoom in the scale for displaying subtleties. At every rule amount level, FastContext outperforms the other two.

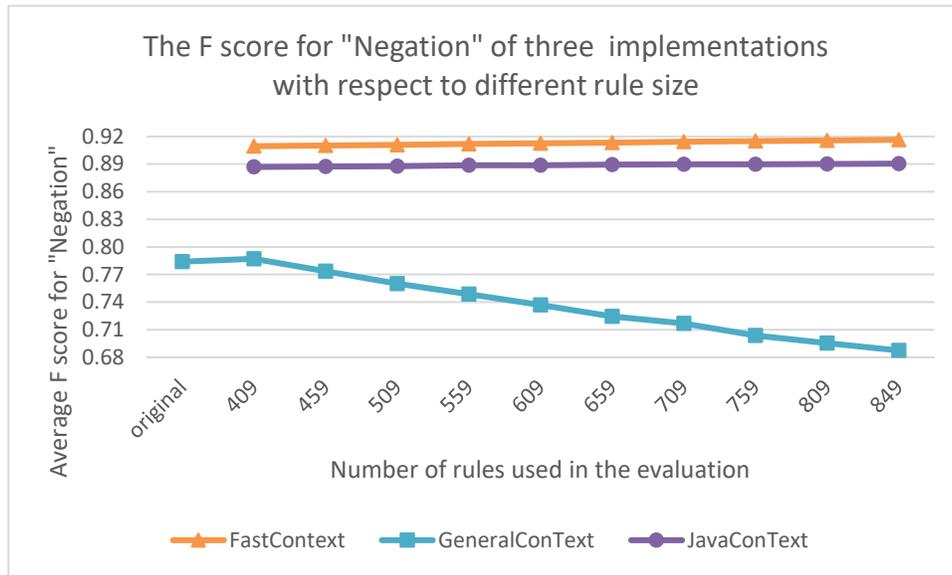

Figure 2. The average F scores for "Negated" detection of three implementations with respect to different rule amount (F score of JavaConText using native rule set is not plotted to zoom in the scale)

Figure 3 compares the F scores of the "Possible" detection between FastContext and JavaConText. The plot indicates that the FastContext's accuracy of "Possible" detection effectively improves as the rule amount grows, while JavaConText is barely effective (F scores below 0.001).

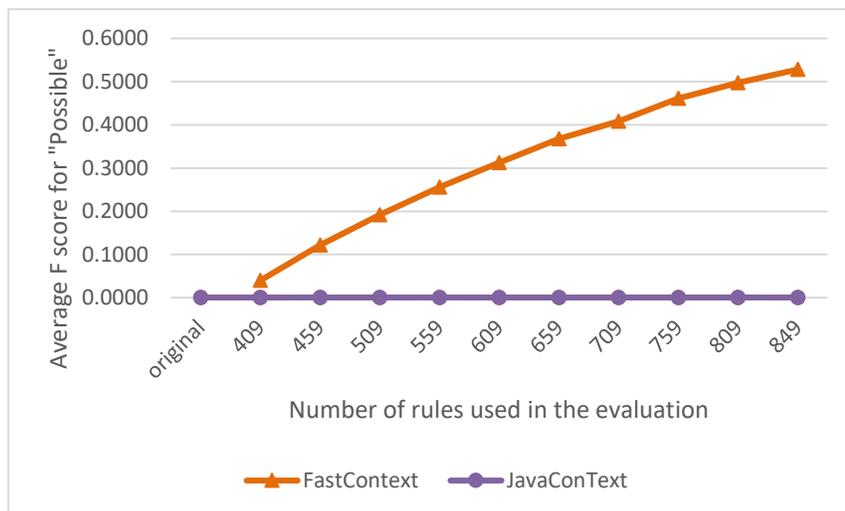

Figure 3. The average F scores for "Possible" detection of FastContext and JavaConText with respect to different rule amount

*Comparing the accuracy of experiencer detection*

In "Other" detection, FastContext exhibits superior performance against JavaConText and GeneralConText. In Figure 4, FastContext shows steeper slope than the other two, indicating that FastContext uses the new added rules more efficiently. GeneralConText experiences a significant accuracy drop when adopting the initial 409 rules. JavaConText shows a slight accuracy drop while the rule increases.

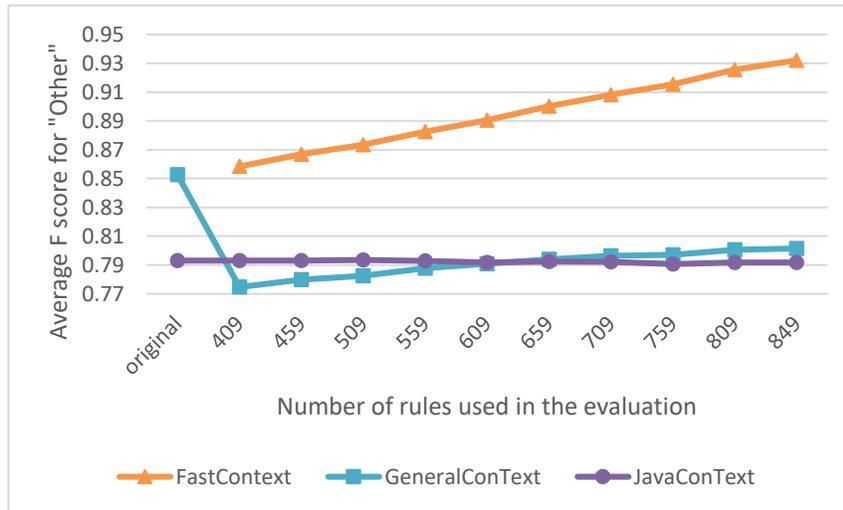

Figure 4. The average F scores for detecting "Other" among three implementations with respect to different rule amount

## DISCUSSION

The FastContext performs up to 800 times faster than GeneralContext and 300 times faster than JavaConText. This speed improvement is crucial when processing large corpora. It saves both time and financial cost. With the accumulation of rules over time, the speed advantage of FastContext will be even more remarkable.

Importantly, FastContext does not achieve accelerated performance by sacrificing accuracy. On the contrary, FastContext improves accuracy by avoiding the shortcomings of JavaConText and GeneralConText, such as non-exhaustive search for the most suitable rules (both of them), non-discriminative to the context clues' directions (GeneralConText) and insufficient support for all types of triggers (GeneralConText).

The results demonstrate the validity of our time complexity estimation. JavaConText and GeneralConText have processing times that increase with the size of the rule set, while FastContext does not. There is a minor non-linear processing time increase, but that could be due to the unpredictability of CPU frequency fluctuations. Because FastContext runs so fast (even 200 runs only take less than 30 seconds), this makes the running time measurement more sensitive to hardware performance fluctuations, especially when compared with the running time of the other two implementations (which take over an hour).

## LIMITATIONS

This study is limited by the evaluation set which is highly imbalanced for experiencer: only 214 out of 16750 test cases are annotated as "Other," which reflects the nature of clinical records in real world, where the focus is primarily on the patient. Nevertheless, we avoid improper evaluation by computing only the F score of "Other" detection. We also could not compare accuracy for temporality.

## CONCLUSION

FastContext implements ConText algorithm through a more effective and efficient approach. In our evaluation, FastContext outperforms JavaConText and GeneralConText in term of both speed and accuracy. In addition, FastContext's speed performance is potentially unaffected by increasing rule amount. Its accuracy performance does not depend on the order of rules, which makes adding customized rules easy.

## ACKNOWLEDGMENTS

Special thanks for Dr. Olga Patterson and Thomas Ginter for sharing their ConText rules for this study.